# Game Agent Driven by Free-Form Text Command: Using LLM-based Code Generation and Behavior Branch

Ray Ito[*1]   Junichiro Takahashi[*2]

[*1*2] University of Tokyo

Several attempts have been made to implement text command control for game agents. However, current technologies are limited to processing predefined format commands. This paper proposes a pioneering text command control system for a game agent that can understand natural language commands expressed in free-form. The proposed system uses a large language model (LLM) for code generation to interpret and transform natural language commands into behavior branch, a proposed knowledge expression based on behavior trees, which facilitates execution by the game agent. This study conducted empirical validation within a game environment that simulates a Pokémon game and involved multiple participants. The results confirmed the system's ability to understand and carry out natural language commands, representing a noteworthy in the realm of real-time language interactive game agents.

## 1. Introduction

The Pokémon franchise began with the release of the first Pokémon games, Pokémon Red and Green, in 1996. In these games, players select four moves for their Pokémon by pressing buttons during battles. However, the subsequent launch of the Pokémon animation series in the following year presented an alternative perspective, depicting a world where Pokémon trainers could direct their Pokémon according to their wishes. Cary (2009) noted that this animated series allows Pokémon enthusiasts to vicariously experience the unique significance of themselves as Pokémon trainers, emphasizing the roles of training and commanding these creatures. Our objective is to bring this experience to life by developing a game that enables players to command their game agents using natural language.

Several attempts have been made to implement language command control for game agents. The first attempt was a game 'Hey You, Pikachu!' released by Nintendo in 1998. It employed pattern matching with a limited vocabulary to identify words from the dictionary [Hobonichi, 99]. Yoshida et al. (2021) expanded on this pattern-matching method by incorporating synonyms, thereby broadening the vocabulary for commanding actions. Mehdi et al. (2009) developed a rule-based control system that could grammatically understand natural language commands and convert them into executable formats. Waqar et al. (2021) introduced deep learning for voice control, but it was limited to precepting predefined words. Therefore, the previously developed language understanding system had a rule-based nature, which could not process unhandled words or forms of the player's language commands. The objective of this paper is to develop a gaming experience that allows players to command their game agents using natural language without any limitations on predefined words.

In the field of robotics, researchers have been studying natural language command control. With the advancements in large language models (LLMs), we are now approaching a stage where systematic and flexible natural language understanding is possible. Liang et al. (2022) demonstrated the use of LLMs to command robots by generating code for executing commands. However, their system is designed for processing independent tasks and is not suitable for continuous actions. Cao and Lee (2023) extended the approach by creating a behavior tree using the same method. However, it is limited to the first command and lacks support for real-time continuous commands.

This paper introduces a new knowledge expression called a behavior branch, which is an endlessly expanding action list similar to a behavior tree. We created a game environment that simulates a Pokémon game. The game agents can be controlled through free-form natural language commands generated by LLM, forming behavior branches in real-time.

The key contributions of this paper include:

1) Enabling game agents to comprehend and execute free-form natural language commands.
2) Introducing a new knowledge expression, the behavior branch, facilitating real-time addition of new commands for continuous actions by the agent. Thus, the game agent can perform continuous actions.

## 2. Methods

### 2.1 Behavior Branch

The behavior tree, a design framework conceptualized by Dromey (2023), has become prevalent in the gaming industry [Miyake 2015]. This structure dictates decision-making processes and actions for game agents, intended for long-term usage by looping through the same tree throughout the agent's lifetime.

Contact 1: Ray Ito ray51ito@g.ecc.u-tokyo.ac.jp Department of Systems Innovation, Faculty of Engineering, The University of Tokyo

Contact 2: Junichiro Takahashi takahashi-junichiro509@g.ecc.u-tokyo.ac.jp Department of Information and Communication Engineering, The University of Tokyo

Adress 1, 2: 7-3-1 Hongo, Bunkyo-Ku, Tokyo

However, this is not suitable for receiving of real-time commands from the player because this is not designed for real-time expansion.

The introduction of a new knowledge expression, the behavior branch, aims to extend the utility of the traditional behavior tree by enabling real-time expansion. Each node in the behavior branch is designed for short-term usage and intended to be used basically once[*1]. The behavior branch is based on three control structures: sequence, selection and repetition; which the concept of the structured programming is composed of [NATO Science Committee 1970].

The behavior branch is a rooted arborescence[*2] and comprises these three fundamental node types:

1) Action node: Specifies the action type and its parameters. When active, the game agent consistently performs the designated action.

2) Condition node: Contains condition information, which has the role of perception in the decision-making. This node is linked to two subsequent nodes: the next node and the true node. While active, the condition outlined within this node undergoes continuous evaluation. If the condition holds true, the next node activates.

3) Control node: Governs the connection or flow of the behavior branch.

Action nodes and control nodes are connected linearly, with each node linked to one previous and one subsequent node. Condition nodes, however, connect to two nodes, as previously mentioned.

The behavior tree includes two variable nodes: the current node and the active action node. The current node shows where in the behavior branch the game agent is focusing. The current node indicates the game agent's current focus within the behavior branch, while the active action node indicates the action node that the game agent is currently performing. The active action node is determined as the last action node found when moving backwards from the current node. The variables' behavior follows specific rules.

1) Sequence: The current node basically typically transitions to the next node of itself when exists, unless the next node is an action node. In the case of an action node, the current node transition waits until the 'satisfied' property of the active action node is true, indicating that the current action has been performed enough for the player to perceive the action underwent.

2) Selection: The condition nodes between the current node and the active action node are evaluated in every game frame. If a condition is confirmed to be fulfilled, the current node will transition to the true node of the condition node.

---
[*1] The node may be reused if there is repetition.
[*2] Arborescence is defined as a rooted tree graph directed away from the root node [Narsingh 1974].

3) Repetition: Within the control nodes, there is a 'repeat' node that causes the subsequent nodes to repeat for a designated number of times. If the current node has reached the leaf node or 'then' node (mentioned later), the current node will return after the 'repeat' node.

As previously stated, the behavior branches are intended for dynamic expansion. The LLM translator generates a new behavior branch from the language command, and it is connected to the game agent's existing behavior branch according to the following rule:

1) When the root of the adding branch is an action node or a control node; the root node will replace the next node of the current node, and then the current node will be changed to it. This is because the player command is intended to have higher priority than regularly waiting for the queued actions following the sequence rule.

2) When the root is a condition node this simply appended to the last of the branch; since this adding branch is intended to add a condition and not an action. In addition, if this is appended while repetition, the evaluation of this condition will be considered as a loop finishing condition.

3) When the root is 'then' control node which has a nuance of 'after that'; the behavior varies by the added branch state:
   a. If the last node of the added node is an action node, the adding branch will be appended to it, without changing the current node immediately
   b. If the last node is a condition node, the adding branch will be appended to the true nodes of the condition.
   c. If the added branch is currently repeating, the added branch will be executed after the end of the repetition.

**Table 1**
This shows each of the simplified connection rules.
'A.', 'Ct.', 'Cd.' and 'T' respectively stands for 'Action', 'Control', and 'Condition' and 'Then control node.'

| Head of Adding Branch | Tail of Added Branch | Connection Rule |
|---|---|---|
| A. or Ct. | Any | Switch immediately |
| T → A. or Ct. | A. or Ct. | Append as next |
| | Cd. | Append as true node of Cd. |
| | (Repeating) | Append as next and execute after repetition finished |
| Cd. | A. or Ct. | Append as next |
| | Cd. | Append as next |
| | (Repeating) | Append as next as ending condition |
| T → Cd. | A. or Ct. | Append as next |
| | Cd. | Append as true node of Cd. |
| | (Repeating) | Append as next and execute after repetition finished |

## 2.2 Game Environment

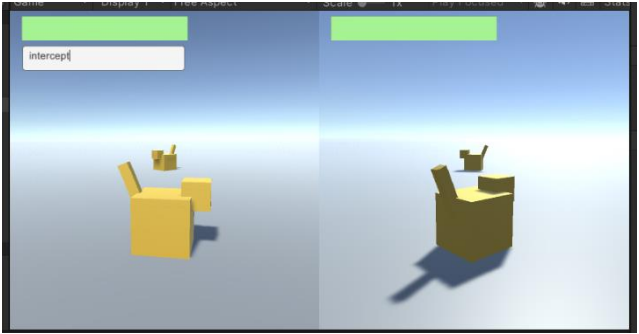

**Fig 2.** The screen of the game environment. The left half represents the player's point of view, while the right half represents the opponent agent's.

To aid in participant familiarization, we developed a simulation environment replicating the dynamics of "PokéPark Wii: Pikachu's Adventure[*1]" using the Unity game development platform[*2]. Within this environment, two game agents are situated in a straightforward two-dimensional plane, both tasked with the objective of attacking their opponent until its Health Points (HP) are depleted to zero. The game agents can move in the plane and attack the opponent. The attacks have three types:

- Tackle[*3]: The agent moves in a high speed straightly and hit its body to the opponent.

- Thunderbolt[*3]: The agent emits a sphere bullet to the front, and the opponent gets damage if the bullet hits it.

- Iron Tail[*3]: The agent spins its body, and if the opponent is in the range, it gets damage.

## 2.3 Behavior Branch Generation via LLM

Following the methodology employed by Liang et al. (2022), we utilized a code generation LLM to translate natural language commands into executable code for the agent's actions. Similar to Liang et al. (2022), we devised a prompt template that serves as a guideline for the LLM, instructing it on how to formulate our intended code[*4]. Subsequently, we directed the LLM to generate the corresponding Python list, which creates the behavior branch. The LLM utilized in this endeavor is the Code Llama 34B model from Meta[*5], accessed through the Fireworks.ai API[*6][*7].

This entire process is implemented in Python[*8] and establishes a connection with the Unity game environment through TCP sockets[*9].

## 3. Experiments

Our experiments aimed to assess the ability of the language understanding system to convert diverse language commands to proper behavior branches.

First, we conducted a survey with participants. We explained the concept of this study and then allowed them to send English text commands to the game agent and observe the response. The game agent was driven by prototype nodes design and a minimum prompt. We recorded a log of the participants' inputs. The gathered information of the participants was limited to their age, years of playing games, years of using English for conversation and country of their longest residence. This was intended to examine the diversity of the commands. The commands which is not possible with the combination of the mentioned abilities of the game agent, or invalid expression as English were not taken into account.

After the survey, we randomly divided the recorded commands 50% as 'train data' and remaining as 'remain data.' We implemented the nodes and prompts were necessary for executing the train data, and execute the 'test data' and evaluate the validity of the game agent's action by 'good' or 'bad' expression.

## 4. Results

The survey was conducted with 4 participants. 73 valid commands in total were input, and 73 commands were used for train data and the remaining 36 commands for test data.

The ratio of 'good' result was 86.11%,

The 'bad' results and their additional information are listed below:

- 'Continue to thunderbolt': The thunderbolt was expected to launch continuously, but it only launched once. However, the command 'Keep doing thunderbolt' worked.
- 'Escape': It was expected for the agent to run away from the enemy, but it intercepted. However, the command 'Escape from opponent' worked.
- 'Attack to the enemy': It was expected to face the enemy and attack it, but it didn't rotate and the attack missed.
- 'The same action': There was no node capable of perceiving the previous action, making it theoretically impossible.
- 'Go behind the opponent': The agent was facing away from the enemy. However, it is important to note that a subsequent attempt at the same action was successful.

In addition, there were no system error. All commands were converted to valid behavior branches.

## 5. Conclusion

The result demonstrated a satisfying ability of the proposed system to convert free-form natural language commands into

---

[*1] © 2009 Pokémon © 1995-2009 Nintendo / Creatures Inc. / GAME FREAK Inc.
[*2] Version 2022.3.15f1 used.
[*3] These are based on Pikachu's action from "PokéPark Wii: Pikachu's Adventure[*1]"
[*4] This is widely known as "few-shot learning"
[*5] https://about.fb.com/news/2023/08/code-llama-ai-for-coding/
[*6] https://fireworks.ai/models/fireworks/llama-v2-34b-code
[*7] We compared each code models in Fireworks.ai, Replicate and Together.ai, and concluded this provider and model had the best balance of latency and generation performance.
[*8] Version 3.9.18 used
[*9] This open-source library used: https://github.com/konbraphat51/UnityPythonConnectionModules

proposed knowledge expressions, and they were successfully executed by the game agent dynamically in real-time.

While the 'good' ratio itself was decent, it can be concluded that the behavior is practical if the commands were specific if we observe the 'bad' commands. All the text commands were able to be expressed theoretically through the behavior branch, indicating a need to improve the language translation. In other words, the response could be enhanced through further prompt engineering, adding more condition nodes or improving the LLM itself.

Thus, this study represents a significant advancement in game agent control through the use of diverse text or voice commands, and highly worth pursuing further. In addition, the study's practical performance suggests that it may be applicable to actual game systems.

The following contributions are desired for further investigation:

1) Decreasing the I/O latency.

    Under our current environment, the total latency was approximately 2.0 seconds. However, the generally desired latency is under 0.4 seconds, which is known as the Doherty Threshold [Doherty 1982]. To put this study in use, it is required to speed up the LLM inference. As another practical idea, if there is a creative effort in the game distracting from the latency, this study could be available immediately.

2) Connection with the voice recognition.

    We did not use speech recognition here because we were featuring the language understanding in this paper and we wanted to evaluate the system without the noise of the speech recognition errors. As mentioned above, quick (at least less than 0.4 seconds) transcription is desired and high accuracy is also required for LLM to understand.

3) Clarifying the best practice of making the code prompt.

    There is no evidence to support that the prompt used in this paper is the most effective. The optimal method for creating the required prompt should be disclosed.

## 6. Acknowledgements

We would like to express our sincere thanks to Dr. Koya Ihara (Cyber Agents, Inc.) for the helpful, insightful and exciting discussions.